# Learning Diagnostic Policies from Examples by Systematic Search

Valentina Bayer-Zubek
School of Electrical Engineering and Computer Science
Oregon State University
Corvallis, OR 97331-3102
bayer@cs.orst.edu

## Abstract

A *diagnostic policy* specifies what test to perform next, based on the results of previous tests, and when to stop and make a diagnosis. Cost-sensitive diagnostic policies perform tradeoffs between (a) the *costs of tests* and (b) the *costs of misdiagnoses*. An optimal diagnostic policy minimizes the expected total cost. We formalize this diagnosis process as a Markov Decision Process (MDP). We investigate two types of algorithms for solving this MDP: systematic search based on the AO* algorithm and greedy search (particularly the Value of Information method). We investigate the issue of learning the MDP probabilities from examples, but only as they are relevant to the search for good policies. We do not learn nor assume a Bayesian network for the diagnosis process. Regularizers are developed that control overfitting and speed up the search. This research is the first that integrates overfitting prevention into systematic search. The paper has two contributions: it discusses the factors that make systematic search feasible for diagnosis, and it shows experimentally, on benchmark data sets, that systematic search methods produce better diagnostic policies than greedy methods.

## 1 INTRODUCTION

A simplified form of the diagnosis process describes the sequence of tests performed by a diagnostician, culminating with a diagnosis. For example, a physician might ask several questions (e.g, patient's age, symptoms), perform simple measurements (e.g., body mass index, temperature), and order laboratory tests (e.g, glucose, insulin) in order to determine the disease of the patient. In this sequential decision making process, the doctor takes into account probabilities of test outcomes, likelihood of diseases, and costs. Both tests and misdiagnoses incur costs. Some tests are cheaper than others, and incorrect diagnoses may incur different costs (for example, declaring a sick patient to be healthy can be more expensive than declaring a healthy patient to be sick).

This paper studies the problem of learning diagnostic policies from data, with the goal of minimizing expected total costs of tests and misdiagnoses. We assume that the training examples record all test results and diagnoses, and that test costs and misdiagnosis costs are given. Because of the costs involved in collecting such training examples, we assume that the training data sets are relatively small.

Our model of diagnosis makes the following assumptions: each test is a pure observation action, so it does not change the patient; tests are performed one-at-a time, and the results are available before the next decision is made; a test need not be repeated, since it returns the same result; tests have discrete values.

Unlike other work on test selection for diagnosis [9, 17, 4], we do not assume a Bayesian network or influence diagram; instead we directly learn a diagnostic policy from the data. The problem of learning diagnostic policies is related to cost-sensitive learning, test sequencing and troubleshooting. Previous work in supervised learning either ignored all costs or considered only attribute costs or only misclassification costs. More recently, both types of costs were investigated by Turney [16], who used genetic search to learn greedy policies, and by Greiner et al. [6], who provided a theoretical algorithm for learning policies with at most a constant number of tests, assuming enough training examples are available to guarantee close-to-optimal performance of these policies. The test sequencing problem [13] deterministically identifies faulty states while minimizing expected test costs. In troubleshooting [8], a system needs to be restored to a functioning state, using pure observations and repair actions.

We formulate the diagnostic learning problem as a Markov Decision Process (MDP) in Section 2. Section 3 shows how to solve the MDP using the systematic search AO* algorithm; it also describes greedy search. Section 4 attacks the issue of learning the MDP model. We propose integrating the learning of probabilities into the search for diagnostic policies. Sections 5 and 6 introduce several regularization methods that reduce the risk of overfitting; some of them also prune the search space. Sections 7 and 8 describe the experiments, and compare the systematic and greedy search algorithms on real-world data sets. Section 9 presents the conclusions and future work.

## 2 DIAGNOSIS FORMALIZED AS A MARKOV DECISION PROCESS

The diagnosis process is a sequential decision making process, so it can be modeled as an MDP [2]. We first describe the actions of the MDP, then the states, and finally the transition probabilities and the expected costs. All costs are positive.

There are $N$ tests and $K$ diagnoses. Test $x_n$ returns the value of attribute $x_n$, and diagnosis action $f_k$ predicts that the correct diagnosis $y$ of an example is $k$. An action (test or diagnosis) is denoted by $a$.

The states $s$ correspond to all possible combinations of measured attributes. For example, state $\{BMI = large, Insulin = low\}$ records the value "large" for Body Mass Index and the value "low" for Insulin. Each training example provides evidence for the reachability of $2^N$ states. With our assumptions, the joint distribution $P(x_1, \ldots, x_n, y)$ is order independent, therefore our state representation has the Markov property. In the start state $s_0 = \{\}$ no attributes were measured. The terminal state is entered once a diagnosis is made. We assume that states that do not appear in the training data have zero probability.

Test action $x_n$ executed in state $s$ will transition to state $s' = s \cup \{x_n = v_n\}$, where $v_n$ is one of the observed values of $x_n$. The probability of this transition is $P_{tr}(s'|s, x_n) = P(x_n = v_n|s)$, and the expected cost is $C(x_n)$, which is the cost of test $x_n$.

Let $MC(f_k, y)$ be the misdiagnosis cost of diagnosing disease $k$ when the correct diagnosis is $y$. The cost of diagnosis $f_k$ is an expectation over the correct diagnoses $y$, taking the value $MC(f_k, y)$ with probability $P(y|s)$, which is the probability that the correct diagnosis is $y$ given the current state $s$. We write $C(s, f_k) = \sum_y P(y|s) \cdot MC(f_k, y)$.

Formally, a policy $\pi$ for an MDP maps states into actions. For a given start state, *a diagnostic policy takes the form of a decision tree*, each internal node specifying a test, and each leaf specifying a diagnosis (see Figure 1). The value of a policy, $V^\pi$, is the expected total cost of following the policy. Note that changing the order of the tests in a policy changes its value function. Solving the MDP means finding an *optimal policy* $\pi^*$ that minimizes $V^\pi(s)$ for all states $s$. Its value is called the *optimal value function* $V^*(s)$.

## 3 SEARCHING FOR DIAGNOSTIC POLICIES

In this section, we assume that the probabilities of the MDP model are known. Instead of searching the entire state space, whose number of states is exponential in the number of tests, we consider algorithms that visit only a fraction of this huge space.

### 3.1 SYSTEMATIC SEARCH (AO*)

The MDP corresponding to our problem has a unique start state and no directed cycles, therefore the space of policies can be represented as an AND/OR graph [7]. The AO* algorithm [11] is an efficient method for computing the optimal policy $\pi^*$ in an AND/OR graph. Unlike dynamic programming algorithms, like value iteration and policy iteration [15], AO* does not need to visit every state of the MDP. Instead, it relies on an *admissible heuristic* that searches only the parts of the search space that look promising to finding the optimal policy.

For details on the AO* implementation for the diagnosis problem, and for proofs of theorems, we refer the reader to [1]. Here, we will describe the admissible heuristic and its cutoffs, and will give an overall idea of how AO* works.

An AND/OR graph alternates between OR nodes and AND nodes. An OR node corresponds to a state $s$ in the MDP, and it specifies the choice of an action (either a test or a diagnosis action). An AND node corresponds to a state-action pair $(s, x_n)$, and stores the probabilities $P(x_n = v_n|s)$ for the outcomes of test $x_n$. Note that multiple paths from the root (corresponding to $s_0$) may lead to the same OR node, by changing the order of the tests. Let $A(s)$ be the set of actions executable in state $s$, including not-yet-measured attributes, and all the diagnosis actions.

Our admissible heuristic provides an optimistic estimate, $Q^{opt}(s, x_n)$, of the expected cost of an unexpanded AND node $(s, x_n)$. It performs a one-step lookahead, and it underestimates the costs of the resulting states $s'$ as the minimum over the cost of diagnosis actions and the cost of each attribute not yet measured in $s'$: $Q^{opt}(s, x_n) = C(x_n) +$ $\sum_{s'} P_{tr}(s'|s, x_n) \cdot \min_{a' \in A(s')} C(s', a')$. The idea is that the cost of a policy that measures at least one test exceeds the cost of that test. We proved that the state-action value function $Q^{opt}(s, a)$ and the state value function $V^{opt}(s) \stackrel{\text{def}}{=} \min_{a \in A(s)} Q^{opt}(s, a)$ form an admissible heuristic, that is, they underestimate the optimal costs, $Q^{opt}(s, a) \leq Q^*(s, a)$ and $V^{opt}(s) \leq V^*(s)$. $V^{opt}$ is the value of the optimistic policy $\pi^{opt}$.

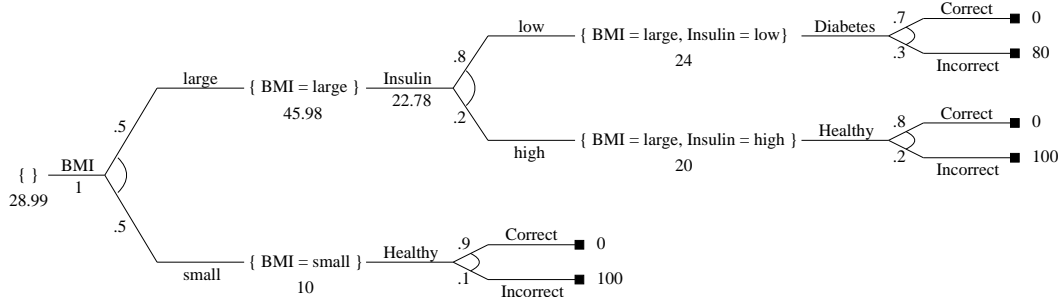

Figure 1: An example of diagnostic policy $\pi$ for diabetes. Body Mass Index (BMI) is tested first. If it is small, a Healthy diagnosis is made. If BMI is large, Insulin is tested before making a diagnosis. We write the costs of tests (BMI and Insulin) underneath them, and the misdiagnosis costs next to the solid squares. Probabilities are written on branches. The values of the states are written below them. The value of the policy, $V^\pi(s_0) = 28.99$, is computed bottom-up by taking expectations of uncertain outcomes and adding test costs.

The admissible heuristic avoids exploring expensive parts of the AND/OR graph. If we computed the optimal state-action value $Q^*(s, a)$, and $Q^*(s, a) < Q^{opt}(s, a')$, then action $a'$ can be pruned from the search space, since it will never be part of the optimal policy. Let us assume that $a$ is a diagnosis action, $a'$ is a test action, and that test costs are large relative to misdiagnosis costs. Then it is likely that the admissible heuristic will produce many cutoffs without expanding expensive actions.

The AO* algorithm repeats the following steps: in the current best optimistic policy (in which not all AND nodes were expanded), it selects an AND node and expands it (that is, it generates its children OR nodes), after which it recomputes the optimistic value function and policy of the revised graph. By definition, a complete policy has diagnosis actions in its leaves. In AO*, a leaf of a complete policy specifies the diagnosis action $f_{best}$ with minimum expected misdiagnosis cost, $f_{best} = \text{argmin}_{f_k} C(s, f_k)$. When AO* converges, the resulting optimistic policy is complete. In fact, this policy is an optimal policy $\pi^*$ of the MDP.

We also introduce the notion of a realistic policy $\pi^{real}$, which is the best complete policy in the graph expanded so far. We compute $\pi^{real}$ by ignoring all unexpanded AND nodes in the current graph; the resulting graph is called the realistic graph. Note that an OR node $s$, where all AND nodes corresponding to remaining tests are currently unexpanded, has $\pi^{real}(s) = f_{best}$. The value of the realistic policy $V^{real}$ is an upper bound on the optimal value function, $V^*(s) \leq V^{real}(s)$. The realistic policy is not necessary for AO* convergence, but it helps us to transform the AO* algorithm into an *anytime algorithm* (where after every iteration we can output a complete, executable policy), and is essential for some of the regularizers.

### 3.2 GREEDY SEARCH

In this section we describe three greedy search algorithms for finding diagnostic policies. Greedy search algorithms perform a limited lookahead search, and once they commit to the choice of a test, that choice is final. As a result, greedy policies are not optimal, but are nevertheless computationally efficient. Instead of growing a graph like AO*, a greedy algorithm builds a single decision tree.

The first greedy method is inspired by the C4.5 algorithm for growing decision trees [14], but it uses Norton's criterion [12]. It selects the test that maximizes the information gain with the diagnoses labels $y$, divided by the cost of the test, $I(x_n; y|s)/C(x_n)$. The information gain is $I(x_n; y|s) = H(y|s) - \sum_{v_n} P(x_n = v_n|s) \cdot H(y|s \cup \{x_n = v_n\})$, where $H(y) = \sum_y -P(y) \log P(y)$ is the Shannon entropy of random variable $y$. If all examples in a node have the same diagnoses, or if all tests have been performed, the greedy search terminates by choosing the most likely diagnosis, $\text{argmax}_y P(y|s)$.

We extend the first greedy method to use misdiagnosis costs in the leaves of the policy. Thus the second greedy method chooses diagnosis actions with the minimum expected cost, $f_{best} = \text{argmin}_{f_k} \sum_y P(y|s) \cdot MC(f_k, y)$.

The last greedy method considers both test costs and misdiagnosis costs at each decision step. The one-step Value of Information (VOI) method first computes the cost of the diagnosis action $f_{best}$ that minimizes expected misdiagnosis costs in state $s$. If all tests have been performed, the method stops by choosing $f_{best}$. If not, for each remaining test $x_n$ it computes the expected cost of performing the test and then choosing diagnosis actions in the resulting states with minimum expected costs, 1-step-LA$(s, x_n) = C(x_n) + \sum_{v_n} P(x_n = v_n|s) \times \left[\min_{f_k} \sum_y P(y|s \cup \{x_n = v_n\}) \cdot MC(f_k, y)\right]$. The best test $x_{best} = \operatorname{argmin}_{x_n}$ 1-step-LA$(s, x_n)$ is selected only when its value of information is positive, $VOI(s, x_{best}) \stackrel{\text{def}}{=} C(s, f_{best}) - $ 1-step-LA$(s, x_{best}) > 0$. Otherwise, it is cheaper to diagnose in $f_{best}$.

# 4 LEARNING PROBABILITIES OF THE MDP MODEL

This section addresses the question of learning the probabilities $P(x_n = v_n|s)$ and $P(y|s)$ of the MDP model. However, not all the probabilities of the MDP model may be required by a search algorithm. Instead of learning the probabilities in a step prior to the search process (e.g., by fitting a probabilistic model to the data, then inferring them from this model), we chose to exploit the task by integrating learning into the search process. This way we only estimate probabilities that are needed for learning good policies.

Each time a search algorithm needs to estimate a probability, the algorithm examines the training data and computes the maximum likelihood estimate. By definition, an example *matches* a state $s$ if it agrees with all the attribute values defining $s$. $P(x_n = v_n|s)$ is estimated as the fraction of training examples matching state $s$ that have $x_n = v_n$. Similarly, $P(y|s)$ is estimated as the fraction of training examples matching state $s$ that have diagnosis $y$.

This simple approach to estimating probabilities often results in *overfitting*, that is, finding policies that perform well (optimally, for AO*) on the training data but perform quite badly on new cases. The following sections describe strategies for reducing overfitting.

# 5 REGULARIZERS FOR SYSTEMATIC SEARCH (AO*)

Both systematic and greedy search algorithms overfit when they grow deep policies whose probabilities are estimated from a small set of training examples. AO* is affected even more by overfitting because it considers many different policies. We first describe strategies for regularizing systematic search. The regularizers change the MDP model. Note that regularized AO* no longer computes the optimal policy on training data.

## 5.1 LAPLACE CORRECTION

Laplace correction avoids extreme probabilities (0 and 1) by adding one fake example to each case. Intuitively, when correcting $P(y|s)$, each diagnosis is given an extra training example. Similarly, when correcting $P(x_n = v_n|s)$, we count one extra example for each value of the test. All probabilities are corrected as the AND/OR graph is grown.

## 5.2 STATISTICAL PRUNING

Our second regularization technique, called *statistical pruning* (SP), reduces the amount of AO* search by pruning actions that are statistically indistinguishable from the current realistic policy.

The SP heuristic is applied in every OR node $s$ whose optimistic policy is selected for expansion. The action $\pi^{opt}(s)$ will be pruned from the graph if a statistical test cannot reject the null hypothesis that $V^{opt}(s) = V^{real}(s)$. The statistical test checks whether $V^{opt}(s)$ falls inside a 95% normal confidence interval around $V^{real}(s)$. If it does, then SP prunes $\pi^{opt}(s)$. In other words, it prefers a complete policy (the realistic policy) to an incomplete one (the optimistic policy). The confidence interval is computed from the total costs of testing and diagnosing all training examples matching state $s$ when processed by $\pi^{real}(s)$.

Recall that the optimal value function $V^*(s)$ is lower-bounded by $V^{opt}(s)$ and upper-bounded by $V^{real}(s)$. If $V^{opt}(s)$ falls inside the confidence interval for $V^{real}(s)$, then $V^*(s)$ will also belong to that confidence interval. Hence, we are at least 95% confident that $V^*(s) = V^{real}(s)$, so the current realistic policy is statistically indistinguishable from the optimal policy. However, subsequent expansions by AO* may change $\pi^{real}$, who could become statistically worse than $\pi^*$.

The SP heuristic is applied as the AND/OR graph is grown. When actions are pruned from the graph, only optimistic updates need to be made, since pruning does not change the realistic graph.

When combining the SP and Laplace regularizers, we center the confidence interval around the Laplace-corrected $V^{real}(s)$, and compute the width of the confidence interval from the total costs of the training examples matching state $s$ when processed by the Laplace-corrected $\pi^{real}(s)$.

## 5.3 EARLY STOPPING

Early stopping employs an internal validation set to decide when to halt AO*. We trained AO* on half of the training data, and used the other half as a validation data. After every iteration, $\pi^{real}$ is evaluated on the validation data. The realistic policy with the lowest total cost on the validation data is remembered, and is returned as the learned policy when the algorithm eventually terminates. When Laplace correction is combined with early stopping, we only correct the probabilities estimated from the subtraining data.

## 5.4 PESSIMISTIC POST-PRUNING BASED ON MISDIAGNOSIS COSTS

This regularizer is inspired by Quinlan's method for pruning decision trees [14]. The idea is to take a policy $\pi$ and the training data, and to produce a pruned policy that exhibits less overfitting. This pruning is applied to the final realistic policy computed by AO$^*$, in a bottom-up traversal of the policy.

Pessimistic post-pruning (PPP) replaces the policy-value of each state, $V^\pi(s)$, by an upper bound $UB(s)$. It starts at the leaves of the policy $\pi$ and computes $UB(s)$ as the upper limit of a 95% normal confidence interval for $C(s, f_{best})$. The confidence interval is computed from the misdiagnosis costs $MC(f_{best}, y)$ of the training examples (with diagnoses $y$) that match state $s$. The upper bound at an internal node is $UB(s) = C(\pi(s)) + \sum_{s'} P_{tr}(s'|s, \pi(s)) \cdot UB(s')$. The action $\pi(s)$ will be pruned, and replaced by the best diagnosis action in $s$, $f_{best}$, if the upper bound on $C(s, f_{best})$ is less than $UB(s)$ for the internal node.

When combining the PPP and Laplace regularizers, we compute the upper bound on $C(s, f_{best})$ by adding one fake training example for each diagnosis. All probabilities were Laplace-corrected as the graph was grown, so $P_{tr}(s'|s, \pi(s))$ used in the computation of $UB(s)$ of internal nodes are already corrected.

## 6 REGULARIZERS FOR GREEDY SEARCH

We now describe regularizers for greedy search.

### 6.1 MINIMUM SUPPORT PRUNING

The first two greedy methods use the minimum support stopping condition of C4.5. Test $x_n$ is eligible for selection only if at least two of its outcomes lead to states that have at least 2 matching training examples.

### 6.2 LAPLACE CORRECTION

Laplace correction is applied to all probabilities computed during greedy search. This does not change the test action with maximum information gain. Laplace correction does not change the most likely diagnosis computed by the first greedy method, but it may change the diagnosis action with the minimum expected cost computed by the second greedy method. For the VOI method, Laplace correction is applied to all probabilities employed in computing $C(s, f_{best})$ and 1-step-LA$(s, x_n)$ as the policy is grown.

Next we describe post-pruning techniques for the greedy policy $\pi$, and discuss how Laplace affects them.

### 6.3 PESSIMISTIC POST-PRUNING BASED ON MISDIAGNOSIS RATES

The first greedy method uses C4.5's standard pessimistic post-pruning. After the tree is grown, in each leaf the pessimistic error is estimated as the upper limit of a 75% confidence interval for the binomial distribution $(n, p)$ plus a continuity correction. $n$ is the number of training examples reaching the leaf node, and $p$ is the error rate committed by the diagnosis action on the training examples at this leaf. An internal node is converted to a leaf node if the sum of its children's pessimistic errors is greater than or equal to the pessimistic error that it would have if it were converted to a leaf node.

Laplace regularization combined with PPP replaces the observed error rate $p$ with its Laplace-corrected version (this is computed by adding one fake example for each diagnosis).

### 6.4 POST-PRUNING BASED ON EXPECTED TOTAL COSTS

The policy $\pi$ grown by the second greedy method is post-pruned based on the expected total cost of diagnosis. An internal node with $\pi(s) = x_n$ is converted into a leaf node, where $\pi(s) = f_{best}$ and $V^\pi(s) = C(s, f_{best})$, if the expected cost of diagnosis, $C(s, f_{best})$, is less than the expected total cost of choosing test $x_n$, $Q^\pi(s, x_n) = C(x_n) + \sum_{s'} P_{tr}(s'|s, x_n) \cdot V^\pi(s')$. When combining this pruning technique with Laplace corrections, all probabilities employed in computing $C(s, f_{best})$ and $Q^\pi(s, x_n)$ were already Laplace-corrected when the policy was grown.

It is interesting to note that this post-pruning based on expected total costs is not necessary for VOI, because pruning is already built-in. Indeed, any internal node $s$ in the VOI policy $\pi$, with $\pi(s) = x_n$, has $Q^\pi(s, x_n) \leq$ 1-step-LA$(s, x_n) < C(s, f_{best})$.

## 7 EXPERIMENTAL STUDIES

We compare the various methods described above, with the goal of finding the best (or the most robust) algorithm.

Table 1: Medical Domains.

| domain | # examples | # tests | (min, max) test cost |
|--------|-----------|---------|---------------------|
| bupa   | 345       | 5       | (7.27, 9.86)        |
| pima   | 768       | 8       | (1, 22.78)          |
| heart  | 297       | 13      | (1, 102.9)          |
| b-can  | 683       | 9       | (1, 1)              |
| spect  | 267       | 22      | (1, 1)              |

## 7.1 EXPERIMENTAL SETUP

The experiments were performed on five medical problems from the UCI repository [3]: Liver disorders (bupa), Pima Indians Diabetes (pima), Cleveland Heart Disease (heart), the original Wisconsin Breast Cancer (b-can), and the SPECT heart database (spect). These data sets describe each patient by a vector of attribute values and a class label. We define a test action that measures the value of each attribute, and a diagnosis action for each class label.

For the bupa, pima, and heart domains, Peter Turney provided the test costs [16]. For the others, we set all test costs to be 1. Assigning misdiagnosis costs is more difficult. We developed a methodology for choosing five different levels of misdiagnosis costs for each domain [1]. The goal was to create an interesting range of misdiagnosis costs relative to test costs, that avoids trivial policies measuring no tests or measuring all tests. Table 1 briefly describes the domains.

We pre-processed the data as follows: we removed all training examples that contained missing attribute values; we merged some of the classes so that only two classes (healthy and sick) remained; we discretized each real-valued attribute into 3 levels (thresholds were chosen to maximize the information gain with the class). For each domain, the transformed data was used to generate 20 random splits into training (two thirds of data) and test sets (one third of data), with sampling stratified by class. Such a split is called a *replica*. Experiments were repeated on each replica to account for random choice of training sets; since the replicas overlap, combining results from different replicas probably underestimate this source of variability.

For domains with many tests, the AND/OR graph constructed by AO* grows very large. To prevent this, we imposed a limit of 100 MB on the total memory for the graph (in practice, this translates into 500 MB). When the memory limit is reached, the current realistic policy is returned as the result of the search. This only happens on the spect domain, for large misdiagnosis costs. In all other cases, the systematic algorithms converge within the memory limit.

The notations for the systematic search algorithms are AO*, SP for AO* with Statistical Pruning, ES for AO* with Early Stopping, and PPP for AO* with Pessimistic Post-Pruning based on misdiagnosis costs. The notations for the greedy search algorithms and their regularizers are Nor, MC-N, and VOI. For all algorithms, the "L" suffix indicates the addition of the Laplace regularizer. For example, MC-N-L denotes the second greedy method using Norton's criterion for selecting tests, and choosing diagnosis actions that minimize expected misdiagnosis costs, along with three regularizers: minimum support pruning, post-pruning based on expected total costs, and Laplace correction.

## 7.2 EVALUATION METHODS

Each algorithm learns a policy on the training set, which we then evaluate on an independent test set. The value of the policy on the test set, $V_{test}$, is the sum of test costs and misdiagnosis cost for each example in the test set, as processed by the policy, divided by the number of examples. To compare learning algorithms, we need to compare their $V_{test}$ values to check if there is a statistically significant difference among them. We used a procedure based on the BDELTACOST bootstrap statistical test [10] to decide whether the policy $\pi_1$ constructed by an algorithm $alg1$ is better than, worse than, or indistinguishable from the policy $\pi_2$ constructed by another algorithm $alg2$.

The original BDELTACOST applies to classifiers that account for misclassification costs but not for attribute costs. We extended the statistical test to diagnostic policies. For each example in the test set, BDELTACOST computes the difference in the total cost of processing it using policy $\pi_1$ and policy $\pi_2$. Then it constructs 1000 bootstrap replicates [5] from the set of cost differences. The means of the bootstrap replicates are sorted in increasing order, and the middle 950 means form a 95% confidence interval for the difference in policies' values. If the confidence interval lies below zero, then $\pi_1$ is better than $\pi_2$ (this is called a *win* for $\pi_1$); if it contains zero, the two policies are *tied*; and if the confidence interval lies above zero, then $\pi_1$ is worse than $\pi_2$ (this is called a *loss* for $\pi_1$).

Let $(wins, ties, losses)$ be the cumulative BDELTACOST results of $alg1$ over $alg2$ on a given domain $D$, across all 5 misdiagnosis cost levels and all 20 replicas. The score of an algorithm is computed using the chess metric, which counts each win as one point, each tie as half a point, and each loss as zero points: $Score(alg1, alg2, D) \stackrel{def}{=} wins + 0.5 \times ties$. The *overall chess score* for an algorithm sums its chess scores against all of the other algorithms: $Score(alg1, D) = \sum_{alg2 \neq alg1} Score(alg1, alg2, D)$. If the total number of "games" played by an algorithm is $Total = wins +$

$ties + losses$, and if all the games were tied, the chess score would be Tie-Score $\stackrel{\text{def}}{=} 0.5 \times Total$. If an algorithm's chess score is greater than the Tie-Score, then the algorithm has more wins than losses.

## 8 RESULTS

We now present the results of the experiments. We first studied the effect of the Laplace regularizer on each algorithm. For each of the seven algorithms with Laplace correction, we computed its chess score with respect to its non-Laplace version, on each domain. The $Total$ number of games an algorithm plays against its non-Laplace version is 100 (5 misdiagnosis cost levels × 20 replicas), so Tie-Score = 50.

Figure 2 shows that on each domain, the Laplace-corrected algorithm scores more wins than losses versus the non-Laplace-corrected algorithm, because each score is greater than Tie-Score. This supports the conclusion that the Laplace correction improves the performance of each algorithm. Some algorithms, such as Nor and AO*, are helped more than others by Laplace.

Since the Laplace regularizer improved each algorithm, we decided to compare only the Laplace-corrected versions of the algorithms to determine which algorithm is the most robust across all five domains. We computed the overall chess score of each Laplace-corrected algorithm against all the other Laplace-corrected algorithms, on each domain. The $Total$ number of games is 600 (an algorithm plays 100 games against each of the 6 "opponents"), so Tie-Score = 300.

Figure 3 shows that the best algorithm (i.e., the one with the largest score) varies depending on the domain: ES-L is best on bupa, VOI-L is best on pima and spect, SP-L is best on heart, and MC-N-L is best on b-can. Therefore no single algorithm is best everywhere. Nor-L is consistently bad on each domain; its score is always below the Tie-Score. This is to be expected, since Nor-L does not use misdiagnosis costs when learning its policy. MC-N-L, which does use misdiagnosis costs, always scores better than Nor-L. The fact that VOI-L is best in two domains is very interesting, because it is an efficient greedy algorithm. Unfortunately, VOI-L obtains the worst score in two other domains: heart and b-can. On average, greedy algorithms run in less than 0.1s, while systematic algorithms have CPU times of at most 1000s.

The only algorithm that has more wins than losses in every domain is SP-L, which combines AO* search, Laplace corrections, and statistical pruning. SP-L always scored among the top three algorithms. Consequently, we recommend it as the most robust algorithm. But in domains with hundreds of tests and diagnosis actions, where SP-L (or any of the systematic search algorithms) is too expensive to run, VOI-L is recommended, since it is the best greedy method.

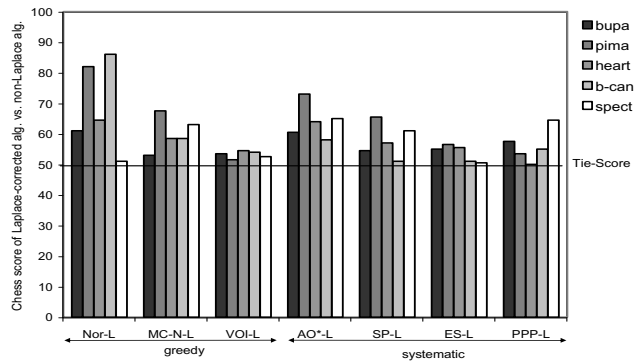

Figure 2: The score of each Laplace-corrected algorithm versus its non-Laplace version, on each domain, is greater than the Tie-Score. Therefore the Laplace version has more wins than losses.

## 9 CONCLUSIONS

This paper addressed the problem of learning diagnostic policies from labeled examples, given both test costs and misdiagnosis costs. The process of diagnosis was formulated as a Markov Decision Problem. We showed how to apply the AO* algorithm to solve this MDP to find an optimal diagnostic policy. We defined an admissible heuristic for AO* that is able to prune large parts of the search space. We also presented three greedy algorithms for finding diagnostic policies.

We integrated the learning of probabilities into the search for good diagnostic policies. To reduce overfitting, we developed four methods for regularizing the AO* search: Laplace corrections, statistical pruning, early stopping, and pessimistic post-pruning. The paper also introduced regularizers for the greedy search algorithms. The algorithms were tested experimentally on five classification problems drawn from the UCI repository. The paper also introduced a methodology for combining the results of multiple training/test replicas into an overall "chess score" for evaluating the learning algorithms.

The experiments showed that all search algorithms were improved by including Laplace corrections when estimating probabilities from the training data. The experiments also showed that the systematic search algorithms were generally more robust than the greedy search algorithms across the five domains. The best greedy algorithm was VOI-L, but although it obtained the best score on two domains, it produced the worst score on two other domains. The most robust learning algorithm was SP-L, combining systematic AO* search with Laplace corrections and statistical pruning.

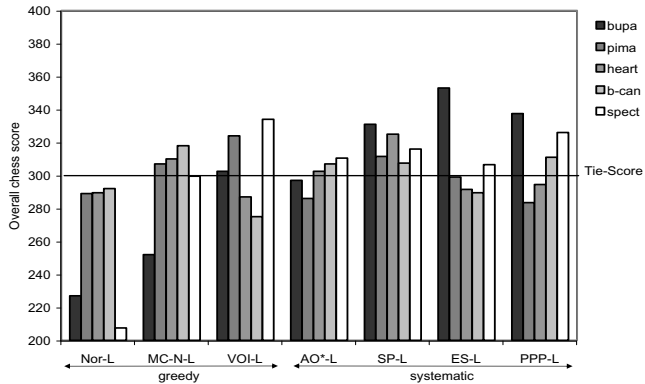

Figure 3: The overall chess score of each Laplace-corrected algorithm, versus all the other Laplace-corrected algorithms. The most robust algorithm is SP-L, being the only one whose score is greater than Tie-Score on every domain.

A surprising conclusion of this paper is that AO$^*$ is computationally feasible when applied to the problem of learning diagnostic policies from training examples. There are three factors that explain this: (a) **The modest amount of training data** limits the number of reachable states in the MDP, and therefore limits the size of the AND/OR graph; the training data has a moderate size because each training example is expensive to collect. (b) **The admissible heuristic** prunes large parts of the search space when test costs are comparable to misdiagnosis costs (which is the case in non-trivial diagnosis problems). (c) **The statistical pruning regularizer** prunes parts of the search space that are unlikely to produce improved policies.

The MDP framework for diagnosis is general enough to handle such extensions as multiple classes and complex costs. The MDP framework needs to be extended to handle treatment actions with side effects, noisy tests, and tests with delayed results. The difficult part for learning is obtaining enough training data for these complex tests. Another challenge is to learn good diagnostic policies from data with missing test results.

**Acknowledgments**

I thank Professor Thomas Dietterich for his guidance.

# References

[1] V. Bayer-Zubek. *Learning Cost-sensitive Diagnostic Policies from Data*. PhD thesis, Department of Computer Science, Oregon State University, Corvallis, http://eecs.oregonstate.edu/library/?call=2003-13, 2003.

[2] V. Bayer-Zubek and T. Dietterich. Pruning improves heuristic search for cost-sensitive learning. In *Proceedings of the Nineteenth International Conference of Machine Learning*, pages 27–35, Sydney, Australia, 2002. Morgan Kaufmann.

[3] C.L. Blake and C.J. Merz. UCI repository of machine learning databases. http:/www.ics.uci.edu/~mlearn/MLRepository.html, 1998.

[4] S. Dittmer and F. Jensen. Myopic value of information in influence diagrams. In *Proceedings of the Thirteenth Conference on Uncertainty in Artificial Intelligence*, pages 142–149, San Francisco, 1997.

[5] B. Efron and R. J. Tibshirani. *An introduction to the bootstrap*. New York: Chapman and Hall, 1993.

[6] R. Greiner, A. J. Grove, and D. Roth. Learning cost-sensitive active classifiers. *Artificial Intelligence*, 139(2):137–174, 2002.

[7] E. Hansen. Solving POMDPs by searching in policy space. In *Proceedings of the Fourteenth Conference on Uncertainty in Artificial Intelligence*, pages 211–219, San Francisco, 1998.

[8] D. Heckerman, J. Breese, and K. Rommelse. Decision-theoretic troubleshooting. *Communications of the ACM*, 38:49–57, 1995.

[9] D. Heckerman, E. Horvitz, and B. Middleton. An approximate nonmyopic computation for value of information. *IEEE Transactions on Pattern Analysis and Machine Intelligence*, 15:292–298, 1993.

[10] D. D. Margineantu and T. Dietterich. Bootstrap methods for the cost-sensitive evaluation of classifiers. In *Proceedings of the Seventeenth International Conference of Machine Learning*, pages 583–590, San Francisco, CA, 2000. Morgan Kaufmann.

[11] N. Nilsson. *Principles of Artificial Intelligence*. Tioga Publishing Co., Palo Alto, CA, 1980.

[12] S. W. Norton. Generating better decision trees. In *Proceedings of the Eleventh International Joint Conference on Artificial Intelligence*, pages 800–805, San Francisco, 1989. Morgan Kaufmann.

[13] K. R. Pattipati and M. G. Alexandridis. Application of heuristic search and information theory to sequential fault diagnosis. *IEEE Transactions on Systems, Man and Cybernetics*, 20(4):872–887, 1990.

[14] J. R. Quinlan. *C4.5: Programs for Machine Learning*. Morgan Kaufmann, San Mateo, California, 1993.

[15] R. S. Sutton and A.G. Barto. *Reinforcement Learning: An Introduction*. MIT Press, Cambrdige, Massachusetts, 1999.

[16] P. D. Turney. Cost-sensitive classification: Empirical evaluation of a hybrid genetic decision tree induction algorithm. *Journal of Artificial Intelligence Research*, 2:369–409, 1995.

[17] L. van der Gaag and M. Wessels. Selective evidence gathering for diagnostic belief networks. *AISB Quarterly*, 86:23–34, 1993.